\documentclass{sig-alternate}

\usepackage{algorithm}
\usepackage{algorithmic}
\usepackage{multirow}

\begin{document}

\conferenceinfo{GECCO'11,} {July 12--16, 2011, Dublin, Ireland.} 
\CopyrightYear{2011} 
\crdata{978-1-4503-0557-0/11/07} 
\clubpenalty=10000 
\widowpenalty = 10000

\title{A Cooperative Coevolutionary Genetic Algorithm for Learning Bayesian Network Structures}

\numberofauthors{1} 
\author{
\alignauthor
Arthur Carvalho\\
\affaddr{David Cheriton School of Computer Science}\\
\affaddr{University of Waterloo}\\
\affaddr{Waterloo, Ontario, Canada}\\
\email{a3carval@cs.uwaterloo.ca}
}

\maketitle

\begin{abstract}
We propose a cooperative coevolutionary genetic algorithm for learning Bayesian network structures from fully observable data sets. Since this problem can be decomposed into two dependent subproblems, that is to find an ordering of the nodes and an optimal connectivity matrix, our algorithm uses two subpopulations, each one representing a subtask. We describe the empirical results obtained with simulations of the Alarm and Insurance networks. We show that our algorithm outperforms the deterministic algorithm K2.
\end{abstract}

\category{\\I.2.6}{Artificial Intelligence}{Learning}



\section{Introduction}

Bayesian networks are graphical models for representing and reasoning under uncertainty~\cite{BN-BOOK}. They provide a means of expressing any joint probability distribution, and in many cases can do so very concisely. The core of a Bayesian network is a directed acyclic graph, whose nodes represent the random variables, and whose edges specify the conditional independence assumptions between the random variables. After construction, a Bayesian network constitutes an efficient tool for performing probabilistic inference.

A Bayesian network can be either constructed ``by hand" or learned from direct empirical observations. Manual network construction is usually impractical since the amount of knowledge required is just too large. An alternative is to learn the network from a set of samples generated from the probability distribution we wish to model.

The task of learning Bayesian network structures from a fully observable data set can be formulated as an optimization problem~\cite{k2,BDeu}. It is proved that this problem  is NP-Hard~\cite{np}. The number of possible structures is super-exponential in the number of nodes~\cite{Rob77}. In this way, structure learning methods usually resort to search heuristics.

An interesting point about the structure learning task is that it can be decomposed into two dependent subtasks: 1) to find an optimal ordering of the nodes, and 2) to find an optimal connectivity matrix. In this paper, we present a method for learning Bayesian network structures from a fully observable data set that explores this idea. It is based on cooperative coevolutionary genetic algorithms where each subtask is represented by a subpopulation.

Besides this introductory section, the rest of this paper is organized as follows. In the next section, we introduce Bayesian networks and the structure learning problem. In Section 3, we present our cooperative coevolutionary  genetic algorithm for learning Bayesian network structures as well as particular choices concerning the genetic operators. In Section 4, we describe the empirical results obtained with simulations of the Alarm and Insurance networks. We show that our solution outperforms the deterministic algorithm K2. In Section 5, we review the literature related to our work. Finally, we conclude in Section 6.

\section{Bayesian Networks}

A Bayesian network is a probabilistic graphical model that represents a joint distribution over a set of random variables, $X_1, \dots, X_n$, by exploiting conditional independence properties of this distribution in order to allow a compact and natural representation~\cite{BN-BOOK}. The core of a Bayesian network is a directed acyclic graph (DAG) $\mathbb{G}$, where nodes represent the random variables and edges correspond to direct influence of one variable on another. Nodes which are not connected represent variables that are conditionally independent of each other. Let $pa(X_i)$ be the set of parents of $X_i$ in $\mathbb{G}$. Thus, there is an edge from each element of $pa(X_i)$ into $X_i$. A Bayesian network decomposes the joint probability distribution $p(X_1, \dots, X_n)$ into a product of conditional probability distributions over each variable given its parents:

\begin{equation}
p(X_1, \dots, X_n)  = \prod_{i = 1}^n p(X_i | pa (X_i)).
\end{equation}

Each node $X_i$ in a Bayesian network is associated with a conditional probability distribution (CPD) that specifies a distribution over the values of $X_i$ given each possible joint assignment of values to the parents of $X_i$. Let $\Theta = (\theta_1, \dots, \theta_n)$ be the vector of parameters that define these conditional probability distributions. In this way, a Bayesian network is fully characterized by the vector $(\mathbb{G}, \Theta )$. Figure 1 shows an example of a classical Bayesian network.

\begin{figure}[h]
\centering
\includegraphics[scale=0.3]{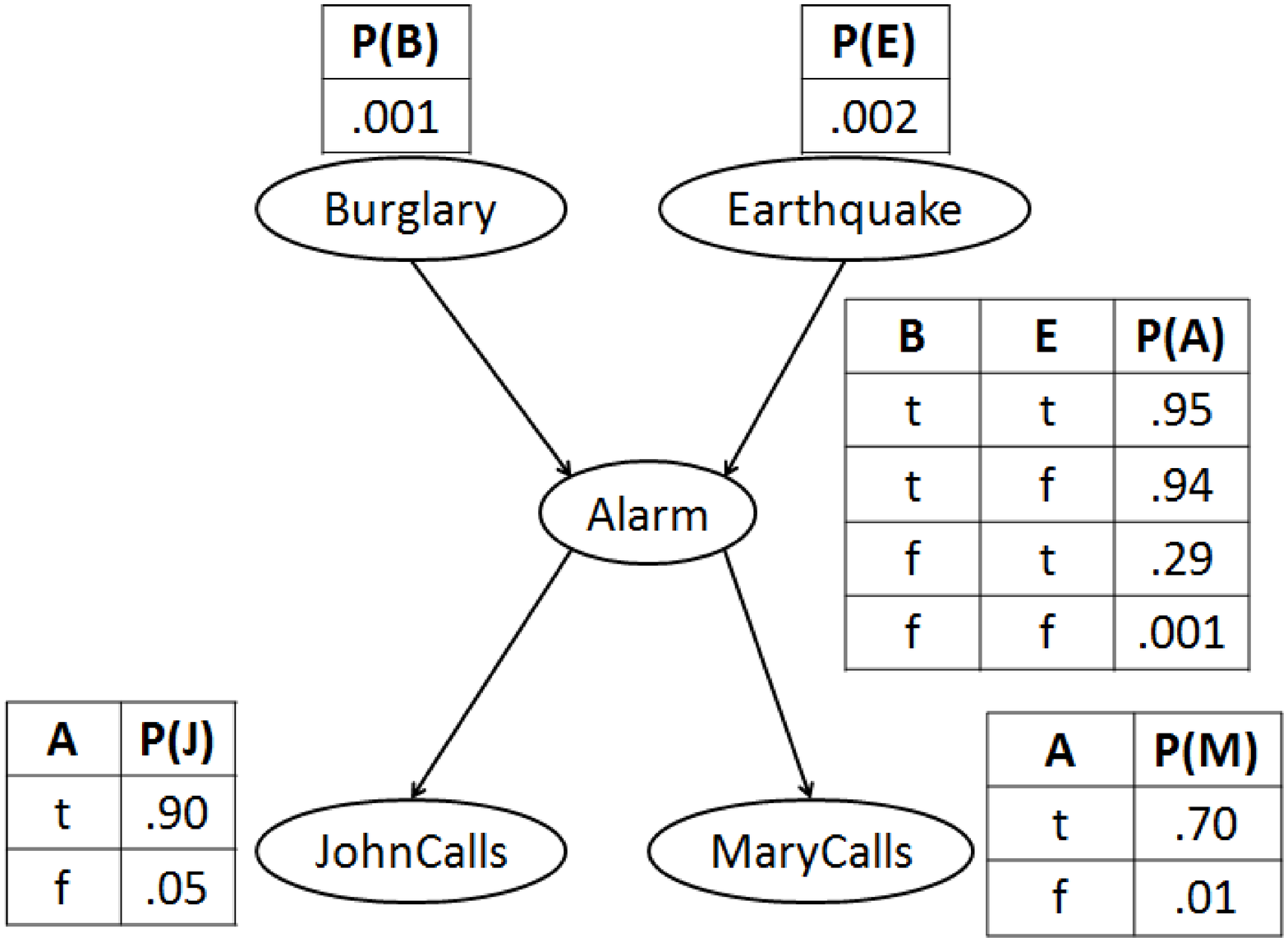}
\caption{Adapted from \cite{BN-BOOK}. Consider a burglar alarm installed at Mr. Holmes' house. It is fairly reliable at detecting a burglary, but also responds on occasion to minor earthquakes. Mr. Holmes has two neighbors, John and Mary, who have promised to call him at work when they hear the alarm. This scenario is represented by the above Bayesian network. Each node is a Boolean random variable. The conditional probability distributions are represented as conditional probability tables. Given the evidence of who has or has not called Mr. Holmes, he is able to estimate the probability of a burglary.}
\end{figure}

A Bayesian network can be either constructed ``by hand" or learned from direct empirical observations. Manual network construction is usually impractical since the amount of knowledge required is just too large. An alternative is to learn the network from a set of samples generated from the distribution we wish to model.

The learning task in Bayesian networks can be separated  into two subtasks: \emph{structure learning}, in which we aim to identify the best topology for a network, and \emph{parameter estimation}, that is to learn the parameters that define the conditional probability distributions for a given network topology. The learning task also depends on the extent of the observability of the data set. We say that the data set is \emph{fully observable} if each training instance contains values for all of the variables in the network. Otherwise, we say that the data set is \emph{partially observable}, \textit{i.e.}, the values of some variables may not be available in all training instances.

In this work, we focus on the structure learning task with fully observable data sets. Roughly speaking, there are three approaches to learning the structure of a Bayesian network~\cite{koller}. The \emph{constrained-based} approach aims to find a network that best explains dependencies and independencies in the data. The \emph{score-based} approach defines a space of potential network structures and a scoring function that measures how well a particular structure fits the observed data.  Finally, the \emph{Bayesian model averaging} approach generates an ensemble of different structures and averages the results provided by them when doing inference. In this paper, we focus on the score-based approach.

A crucial observation about the structure learning task is that it can be decomposed into two dependent subtasks: First, to find an optimal ordering of the nodes where each node $X_i$ can only have node $X_j$ as a parent if node $X_j$ comes before node $X_i$ in the ordering. Second, to find an optimal connectivity  matrix. We explore this idea in our algorithm for learning Bayesian network structures.

\subsection{Bayesian Score Function}

A traditional method for deriving a score function to evaluate Bayesian network structures is based on Bayesian considerations, \textit{i.e.}, whenever we have uncertainty over anything, we should place a distribution over it. In practice, this means to define a \emph{structure prior}, $p(\mathbb{G})$, that puts a prior probability on different graph structures, and a \emph{parameter prior}, $p(\Theta | \mathbb{G})$, that puts a probability on different parameters $\Theta$ given a graph $\mathbb{G}$. Consider a fully observable data set $D$. In a Bayesian score function, we evaluate the posterior probability of a graph $\mathbb{G}$ given the data $D$:

\begin{eqnarray}
\label{Equation:Bayesian_Score}
P(\mathbb{G} | D) &=& \frac{ P(D | \mathbb{G}) P(\mathbb{G})}{P(D)}\\
       &\propto&  P(D|\mathbb{G})P(\mathbb{G}) \nonumber
\end{eqnarray}

\noindent where the equality follows from Bayes' theorem, $P(D| \mathbb{G}) = \int_{\theta} P(D|\mathbb{G}, \theta)P(\theta|\mathbb{G})d\theta$, \textit{i.e.}, it is the marginal likelihood that averages the probability of the data $D$ over all possible parameter assignments to $\mathbb{G}$. The denominator in Equation~\ref{Equation:Bayesian_Score} is simply a normalizing factor that does not help distinguish between different structures. Thus, we can disregard it.

Once the distributions $P(\mathbb{G})$ and $P(\theta|\mathbb{G})$ are specified and the data $D$ is given, structure learning amounts to finding the graph $\mathbb{G}$ that maximizes $P(D|\mathbb{G})P(\mathbb{G})$. The ability to ascribe a prior over structures gives us a way of preferring some structures over others. For example, we can penalize dense structures more than sparse ones. Koller and Friedman \cite{koller} show that although this prior is indeed a bias towards certain structures, in fact, it plays a relatively minor role in Equation~\ref{Equation:Bayesian_Score}. For this reason, it is often used a uniform prior over structures. In this way, the structure learning task is reduced to finding the structure $\mathbb{G}$ with maximum likelihood $P(D|\mathbb{G})$, \textit{i.e.}, a structure that makes the observed data as likely  as possible. If we use a Dirichlet parameter prior for all parameters in the network, then the likelihood $P(D|\mathbb{G})$ can be obtained in closed form~\cite{k2,BDeu}:

\begin{equation}
\label{Equation:BDeu}
P(D|\mathbb{G}) = \prod_{i=1}^n \prod_{j=1}^{q_i}\frac{\Gamma\left(N^\prime_{ij}\right)}{\Gamma\left(N^\prime_{ij} + N_{ij}\right)} 
\prod_{k=1}^{r_i}\frac{\Gamma\left(N^\prime_{ijk} + N_{ijk}\right)} {\Gamma\left(N^\prime_{ijk}\right)} 
\end{equation}

\noindent where $n$ is the number of variables in the network, $r_i$ is the number of possible values for the variable $X_i$, $q_i$ is the number of possible joint assignment of values to the parents of $X_i$, $N_{ijk}$ is the number of occurrences of configurations of variables and their parents, $N^\prime_{ijk}$ are the hyperparameters of the Dirichlet distribution (prior counts of occurrences of variables and their parents), $N_{ij} = \sum_{k=1}^{r_i}N_{ijk}$, $N^\prime_{ij} = \sum_{k=1}^{r_i}N^\prime_{ijk}$, and $\Gamma$ is the Gamma function, which satisfies $\Gamma(m) = (m-1)!$. We assume that the Dirichlet priors are non-informative, \textit{i.e.}, all the hyperparameters have the same value. For simplicity's sake, we assume that this value is equal to $1$. Equation~\ref{Equation:BDeu} is usually referred as the \emph{BDe score}. In practice, the logarithm of Equation~\ref{Equation:BDeu} is usually used since it is more manageable to be computed numerically.

Thus, the structure learning task can be seen as an optimization problem, where we wish to find the structure $\mathbb{G}$ that maximizes the objective function in Equation~\ref{Equation:BDeu}. Chickering \textit{et al.}~\cite{np} prove that this problem is NP-Hard. Robinson~\cite{Rob77} shows that $r(n$), the number of different structures for a network with $n$ nodes, is given by the  recursive formula:

\begin{equation*}
r(n) = \sum_{k=1}^{n}(-1)^{k+1}\binom{n}{k}2^{k(n-k)}r(n-k) = n^{2^{O(n)}}
\end{equation*}

\noindent \textit{i.e.}, the number of DAGs as a function of the number of nodes, $r(n)$, is super-exponential in $n$. For illustration, when $n = 6$, there are $3,781,503$ possible DAGs. When $n = 10$, this number is approximately $4.2\times 10^{18}$. In this way, structure learning methods usually resort to search heuristics.

An interesting property of the BDe score that is often used in order to make this search effective is the \emph{score decomposability}~\cite{koller}. In short, a Bayesian score function $\mathbb{S}(\mathbb{G})$ is decomposable if it can be written as the sum of functions that depend only on one node and its parents, \textit{i.e.}:

\begin{equation*}
\mathbb{S}(\mathbb{G}) = \sum_{i=1}^n f(X_i, pa(X_i)),
\end{equation*}

\noindent where $f$ is a local score function. The major benefit of this property is that a local change in the structure of a DAG (such as adding or removing an edge) does not alter the scores of other parts that remained unchanged. Thus, decomposable score functions may drastically reduce the computational overhead of evaluating different structures.

Another important property of the BDe score is that it is \emph{consistent}. Asymptotically, consistent score functions prefer structures that exactly fit the (in)dependencies in the data~\cite{koller}. This implies that the structure $\mathbb{G}^*$, which is a perfect map of the joint distribution we wish to model, maximizes the score returned by Equation~\ref{Equation:BDeu} when the number of training instances goes to infinity.

\subsection{Equivalence Classes of Bayesian Networks}

Different Bayesian network structures are \emph{equivalent} when they encode the same set of conditional independence assertions~\cite{equivalence,equivalence2}. Consequently, we cannot distinguish between equivalent networks based on observed independencies. This suggests that we should not expect to distinguish between equivalent networks based on observed data cases. We say that a Bayesian score function $\mathbb{S}$ is \emph{equivalent} when for all equivalent networks $\mathbb{G}$ and $\mathbb{G}^\prime$ we have that $\mathbb{S}(\mathbb{G}) = \mathbb{S}(\mathbb{G}^\prime)$. In other words, \emph{score equivalence} implies that all networks in the same equivalence class have the same score.

Heckerman \textit{et al.}~\cite{BDeu} show that the BDe score is equivalent. Furthermore, if we insist on using Dirichlet priors and having the score decomposability property, then the only way to satisfy score equivalence is by using the BDe score. This notion of equivalence plays an important role in the task of learning Bayesian network structures. Since there may be a variety of structures that are equally optimal, methods that perform a multi-directional search (\textit{e.g.}, genetic algorithms) are highly suitable to be applied to this task.

\section{Genetic Algorithms}

A genetic algorithm (GA) is a search heuristic inspired by the theory of evolution. Given a population of individuals, \textit{i.e.}, potential solutions to an optimization problem, each one encoded using a chromosome-like data structure (character strings), GAs work by applying dedicated operators inspired by the natural evolution  process (\textit{e.g.}, selection, crossover, and mutation) to these individuals until a termination criterion is satisfied. The purpose of using a GA is to find the individual from the search space (population) with the best ``genetic material". The quality of an individual is measured with an objective function, also called the fitness function.

Since the structure learning task can be decomposed into two dependent subtasks, it is natural to consider a genetic algorithm that evolves two different subpopulations (species) in a cooperative way, where each individual species represents part of a complete solution. The major issues with this approach are how to represent such individual species and how to apportion credit for them given the fitness of a complete solution. A class of GAs that deals with these issues is called cooperative coevolutionary genetic algorithms.

\subsection{Cooperative Coevolutionary Genetic Algorithms}

Coevolution refers to a reciprocal evolutionary change between species that interact with each other. The most common types of coevolution are based either on competition or on cooperation. They differ from each other in the way that the fitness of an individual species is calculated. In the competitive coevolution, this fitness is the result of a direct competition between different species. In the cooperative coevolution, the fitness of an individual species is resulting from its collaboration with other species.

Our solution to the task of learning Bayesian network structures from a fully observable data set is based on the cooperative coevolutionary genetic algorithm (CCGA) proposed by Potter and De Jong~\cite{CCGA}. CCGA decomposes a problem into a fixed number of subcomponents, each one represented by a different subpopulation. For example, if the solution to an optimization problem consists of the values of $x$ parameters (variables), then a natural decomposition is to maintain $x$ subpopulations, each of which contains competing values for a particular parameter. Thus, CCGA divides a problem into smaller subproblems and solves them in order to solve the original problem.

The evolution of each subpopulation is handled by the standard GA. A complete solution to the original problem is obtained by assembling representative members of each subpopulation. These members are scored based on the fitness of the complete solution in which they participate. Thus, the fitness of a species is computed by estimating how well it ``cooperates" with other species to produce good solutions. Algorithm 1 presents a full description of CCGA.

\begin{algorithm}
\caption{CCGA}
\begin{algorithmic}[1] 
\STATE gen = 0
\FOR{each species $s$} 
\STATE $\mathbf{P}_s(gen)$ = randomly initialized population
\STATE evaluate each individual in $\mathbf{P}_s(gen)$ 
\ENDFOR
\WHILE{termination criterion = false} 
\STATE gen = gen +1
\FOR{each species $s$} 
\STATE select $\mathbf{P}^{\prime}_{s}(gen)$ from $\mathbf{P}_{s}(gen-1)$ based on fitness
\STATE apply crossover and  mutation operators to $\mathbf{P}^{\prime}_{s}(gen)$
\STATE evaluate each individual in $\mathbf{P}^{\prime}_{s}(gen)$
\STATE select $\mathbf{P}_s(gen)$ from $\mathbf{P}^{\prime}_{s}(gen)$ and $\mathbf{P}_{s}(gen-1)$
\ENDFOR
\ENDWHILE
\end{algorithmic}
\end{algorithm}

The algorithm starts by initializing a separate subpopulation, $\mathbf{P}_s$, for each species $s$. The initial fitness of each subpopulation member is computed by combining it with a random individual from each of the other subpopulations, and applying the fitness function to the resulting solution. Thereafter, each subpopulation is coevolved using the canonical GA. For evaluating individual species, each subpopulation member is combined with both the best known individuals and with a random selection of individuals from the other subpopulations. The fitness function is then applied to the two resulting solutions, and the highest value is returned as that subpopulation member's fitness\footnote{The original CCGA uses a more greedy credit assignment in which each individual in a subpopulation is only combined with the best known individuals from the other subpopulations. The CCGA used in this paper is actually called CCGA-2 by Potter and De Jong~\cite{CCGA}. It is a variant of the original CCGA that performs better when there is interdependence between the species.}. In this work, we use the BDe score (Equation~\ref{Equation:BDeu}) as the fitness function. In the rest  of this section, we discuss how to represent a DAG using multiple subpopulations as well as particular choices concerning the genetic operators.

\subsection{Representation}

As discussed in Section 2, the structure learning task  can be decomposed into two dependent subtasks, that is to find an optimal ordering of the nodes and an optimal connectivity matrix. We deal with this problem by using CCGA with two subpopulations, each one representing a subtask.

The first subpopulation, henceforth called the \emph{permutation subpopulation}, represents the ordering task. Its individuals are represented by permutations of the random variables $X_1, X_2, \dots, X_n$. We implicitly assume that the parents of the node at position $i$ come before this node in the ordering, \textit{i.e.}, their positions are between $1$ and $i-1$. In this way, the node at position $1$ can be a parent of $n-1$ nodes, the node at position $2$ can be a parent of $n-2$ nodes, and so on. Under this representation, a fully connected Bayesian network has $\frac{(n-1)n}{2}$ edges, which is the sum of the terms of the finite arithmetic progression $n-1, n-2, \dots, 1$.

The second subpopulation, henceforth called the \emph{binary subpopulation}, represents the task of finding a connectivity matrix. Its individuals are represented by binary vectors of length $\frac{(n-1)n}{2}$. In detail, let $c_{i,j}$ be defined as follows:

\begin{displaymath}
c_{i,j}= \left\{ \begin{array}{ll}
1 & \textrm{if the node at position $i$ is a parent of the node}\\
  & \textrm{at position $j$,}\\
0 & \textrm{otherwise}.
\end{array} \right.
\end{displaymath}

We represent an individual from the binary subpopulation by using the following binary string:

\begin{displaymath}
c_{1,2}, c_{1,3}, \dots, c_{1,n}, c_{2,3}, \dots, c_{2,n}, \dots, c_{n-1,n}
\end{displaymath}

Another way to see an individual from the binary subpopulation is as a strictly upper triangular matrix, that is a matrix having 0s along the diagonal as well as below it. Table 1 shows an example of such perspective for $n=4$.
\begin{table}[H]
\centering
\caption{An individual from the binary subpopulation as a strictly upper triangular matrix.}
\begin{tabular}{|c|c|c|c|c|} \hline
  & \textbf{1} & \textbf{2}         & \textbf{3}         & \textbf{4} \\ \hline
\textbf{1} & 0 & $c_{1,2}$ & $c_{1,3}$ & $c_{1,4}$ \\ \hline
\textbf{2} & 0 & 0         & $c_{2,3}$ & $c_{2,4}$ \\ \hline
\textbf{3} & 0 & 0         & 0         & $c_{3,4}$ \\ \hline
\textbf{4} & 0 & 0         & 0         & 0 \\\hline
\end{tabular}
\end{table}

\subsubsection{Complete Solutions}

Let $I_p$ and $I_b$ be individuals respectively from the permutation and the binary subpopulations. Further, let $I_c$ be the complete solution (a DAG) created by combining $I_p$ and $I_b$. The construction of $I_c$ is done by taking one allele value at a time from $I_p$, say the one at position $i$, followed by $n-i$ consecutive allele values from $I_b$, where $i$ goes from $1$ to $n-1$. In other words, we select a node at a time from $I_p$ together with its out-edges from $I_b$. Thus, $I_c$'s length is equal to $n +\frac{(n-1)n}{2}$. Algorithm 2 describes this process of combining species.  Figure 2 shows an example of a Bayesian network structure and how it is encoded using our representation.  
\begin{algorithm}[H]
\caption{Combining species to create a DAG}
\begin{algorithmic}[1] 
\REQUIRE {Two individuals: $I_p$ and $I_b$}
\ENSURE {A complete solution: $I_c$}
\STATE $\textrm{index}_{\textrm{bin}}$ = 1    
\STATE $\textrm{index}_{\textrm{comp}}$ = 1
\FOR{$ i = 1 \textbf{\textrm{ to }} n-1$} 
\STATE $I_c[\textrm{index}_{\textrm{comp}}] = I_p(i)$
\STATE $\textrm{index}_{\textrm{comp}} = \textrm{index}_{\textrm{comp}}+ 1$
\STATE $I_c[\textrm{index}_{\textrm{comp}} \textbf{\textrm{ to }} 
\textrm{index}_{\textrm{comp}} +  n-i-1 ] = $\\
       $I_b[\textrm{index}_{\textrm{bin}} \textbf{\textrm{ to }} \textrm{index}_{\textrm{bin}} +  n-i-1 ]$
\STATE $\textrm{index}_{\textrm{comp}} = \textrm{index}_{\textrm{comp}}  + n-i$
\STATE $\textrm{index}_{\textrm{bin}} = \textrm{index}_{\textrm{bin}} + n-i$
\ENDFOR
\STATE $I_c[\textrm{index}_{\textrm{comp}}] = I_p(n)$
\end{algorithmic}
\end{algorithm}

\begin{figure}[H]
\centering
\label{representation}
\includegraphics[scale=0.19]{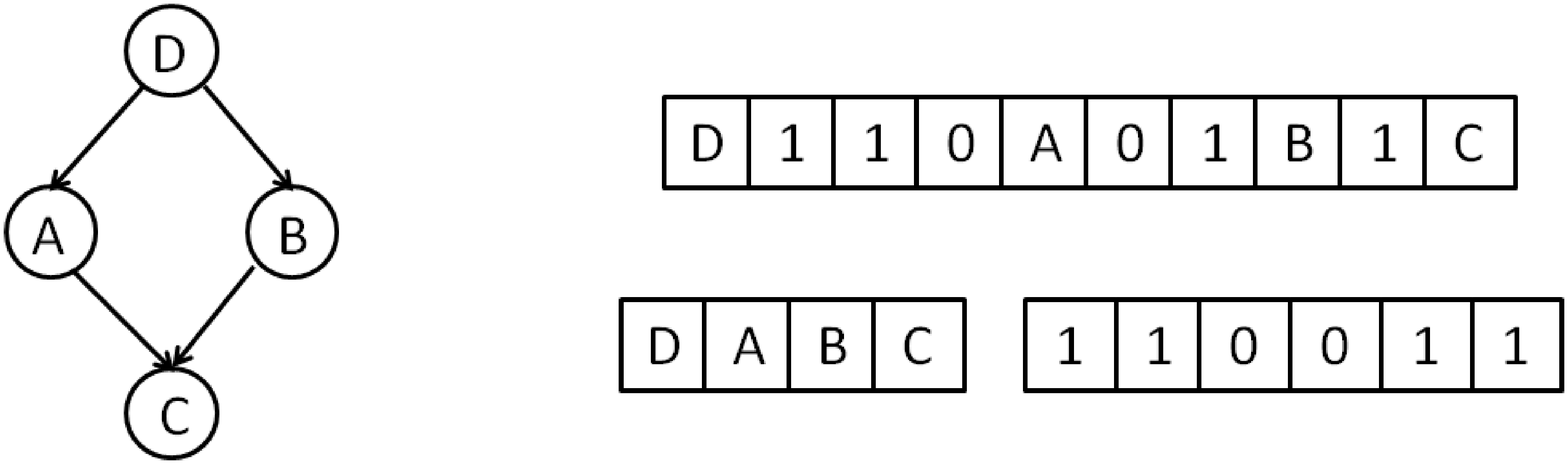}
\caption{(Left) An example of a Bayesian network structure. (Top Right) The structure on the left encoded using the proposed representation. The ancestors always come before the descendants. A numeric cell represents an out-edge of the closest non-numeric cell on the left. For example, the first numeric cell after the node $D$ represents the edge from $D$ to the first non-numeric cell after $D$, that is $A$. The second numeric cell represents the edge from $D$ to $B$, and so on. The value $1$ means that an edge exists between the two nodes, while $0$ means the opposite. (Bottom Right) The individual species that compose the complete solution.}
\end{figure}

Our representation always produces legal network structures since the acyclicity constraint is never violated, \textit{i.e.}, the underlying graphs will never have cycles. This follows from the facts that the individuals from the binary subpopulation are represented by strictly upper triangular matrices and that we implicitly specify node orderings. Thus, our representation is \emph{correct} and, consequently, we do not need to use repair operators to convert invalid DAGs into valid ones (for example, see~\cite{two_evol,Larranaga94structurelearning}). Furthermore, we do not need to detect cycles, thus avoiding extra  computations. Our representation is also \emph{complete} because every single Bayesian network structure can be represented in it.

\subsection{Initialization}

Each individual from the binary subpopulation is semi-randomly initialized in such a way that every single node in a complete solution has only one parent, except the root. For the permutation subpopulation, each individual is randomly initialized without extra procedures. Thus, the initial DAGs are valid and sparse graphs.

\subsection{Selection}

For both subpopulations, we use the \emph{tournament selection} operator. In short, each individual in a subpopulation is duplicated, the individuals are then paired up, and finally the best individual in each pair is selected to produce the offspring. Thus, each individual participate in exactly two tournaments. This operator has better or equivalent convergence and computational time complexity properties than other selection operators that exist in the GA literature~\cite{Goldberg91acomparative}.

\subsection{Crossover}

The selected individuals  are paired up again, and with probability $p_c$ each pair generates two new individuals. Otherwise, the offspring are exact copies of the parents. For the binary subpopulation, we use the traditional \emph{two-point crossover}, where each parent is randomly broken into three segments, and then the offspring are created by taking alternative segments from the parents.

For the permutation subpopulation, we need to take into account the fact that the absolute position of a node inside an individual matters because this defines the ancestor-descendant relationships. A crossover operator that tries to preserve as much information as possible about the absolute positions in which elements occur is the \emph{cycle crossover}~\cite{cycle_crossover}. This operator starts by dividing the elements into cycles, as described in Algorithm 3. Thereafter, the offspring are created by selecting alternate cycles from each parent.

It is important to note that neither the two-point crossover nor the cycle crossover violates the acyclicity constraint. Because of this, we say that they are \emph{closed operators}.

\begin{algorithm}[H]
\caption{Cycle Crossover}
\begin{algorithmic}[1] 
\REQUIRE Two permutations: $P_1$ and $P_2$
\ENSURE Cycles
\STATE Look at the first \emph{unused} position in $P_1$ (say $i$)
\STATE Look at the element $x$ in the \emph{same position} in $P_2$
\STATE Go to the position $j$ in $P_1$ that contains the element $x$ 
\STATE Add $x$ to the cycle $C$
\STATE Repeat steps 2 through 4 while $i \neq j$
\STATE Return the cycle $C$
\STATE Repeat steps 1 through 6 while there is a position in $P_1$ that was not used
\end{algorithmic}
\end{algorithm}

\subsection{Mutation}

For promoting diversity, the resulting offspring may undergo some kind of randomized change (mutation). For the binary subpopulation, we use the traditional \emph{bit-flip} mutation, where each gene is flipped (\textit{i.e.}, from 1 to 0 or 0 to 1) with a small probability $p_{mb}$. Intuitively, we are adding (or removing) a particular edge to a DAG with a small probability $p_{mb}$. For the permutation subpopulation, we use the \emph{swap mutation}. With probability $p_{mp}$, this operator selects two genes at random and swaps their allele values. Intuitively, we are randomly changing ancestor-descendant relationships inside an individual. It is interesting to note that both mutation operators are closed operators.

\subsection{Replacement}

In order to preserve and use previously found best individuals in subsequent generations, we use an elitist replacement strategy. For each subpopulation, the current best individual in  $\mathbf{P}_{s}(gen-1)$ is preserved and automatically copied to the next generation, $\mathbf{P}_{s}(gen)$. The rest of  $\mathbf{P}_{s}(gen)$ is composed by the offspring, except the child with the worst fitness. In this way, the statistics of each subpopulation-best solution cannot degrade with generations.

\section{Experiments}

We compared the performance of our method, using the parameters shown in Table 2, with the deterministic algorithm K2~\cite{k2}. The K2 algorithm can be seen as a greedy heuristic. It starts by assuming that each node has no parents. Hence, for a given node, it incrementally adds the parent whose addition most increases the score of the resulting structure. It stops adding parents when the addition of a single parent cannot increase the overall score. This algorithm has a major drawback, which is to require a predefined ordering of the nodes. In our experiments, we used random permutations as input to K2. This algorithm also requires an upper-bound on the number of parents that a node may have. We set this value to ten. This is a reasonable value since none of the structures used in our experiments have a node with ten or more parents. Our implementation of the K2 algorithm is based on the Bayes Net Toolbox~\cite{murphy2001bnt}.

We used two well-known Bayesian networks in our experiments, namely Alarm and Insurance. The Alarm network~\cite{Dataset:alarm} was constructed for monitoring patients in intensive care. It has 37 nodes and 46 edges. Its structure can be seen in Figure 3. The Insurance network~ \cite{Dataset:Insurance} was constructed for evaluating car insurance risks. It contains 27 nodes and 52 edges. Its structure can be seen in Figure 4. For each Bayesian network, we generated three data sets containing, respectively, 1000, 3000, and 5000 instances.

\begin{table}[H]
\centering
\caption{Parameters of CCGA. $\mathbb{E}$ is equal to the maximum number of edges that a Bayesian network structure may have, \textit{i.e.}, $\mathbb{E} = \frac{(n-1)n}{2}$, where $n$ is the number of nodes.}
\begin{tabular}{|c|c|} \hline
\textbf{Parameter} & \textbf{Value} \\ \hline
Number of generations & 250  \\\hline
Population size       & 100  \\\hline
$P_{mb}$              & $1/\mathbb{E}$  \\\hline
$P_{mp}$              & 0.5  \\\hline
$P_c$                 & 0.6  \\\hline
\end{tabular}
\end{table}

\begin{figure}[ht]
\centering
\psfig{file=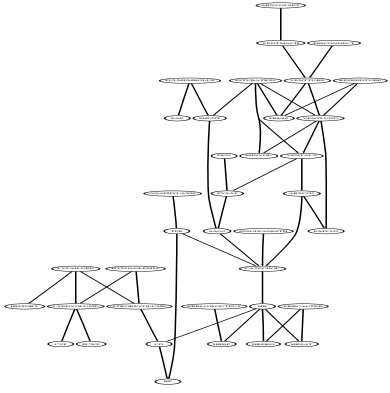, height=2in, width=2in,}
\caption{The structure of the Alarm network.}
\vskip -6pt
\end{figure}

\begin{figure}[ht]
\centering
\psfig{file=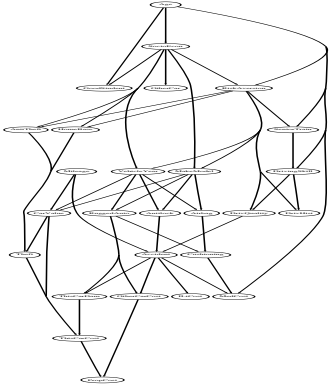, height=2in, width=1.75in,}
\caption{The structure of the Insurance network.}
\vskip -6pt
\end{figure}

\subsection{Results}

For each data set, we ran both algorithms 100 times. In each run, we stored the best structure found by CCGA. The results related to the Alarm network and the Insurance network are respectively shown in Table 3 and 4. Besides standard descriptive statistics, these tables also contain $p$-values from one-tailed t-tests, where the alternative hypothesis is that the mean of the scores resulting from CCGA is greater than the mean of the scores resulting from K2. Below the name of each data set is the score of the original DAG. We note that such scores are not necessarily the highest ones: the data sets being of finite sizes may not represent all the (in)dependencies within the original structure. However, since the BDe score is consistent, the scores of the original structures quickly become the highest ones when the number of samples increases. The convergence of CCGA, averaged over 100 runs, is shown in Figure 5 and 6.

\begin{table*}
\centering
\caption{Results with the Alarm network.}
\begin{tabular}{|c|c|c|c|c|c|c|} \hline
\textbf{Data Set} & \textbf{Algorithm} & \textbf{Average} & \textbf{Standard Deviation} & \textbf{Minimum} & \textbf{Maximum} & \textbf{$p$-value}\\ \hline
Alarm 1000     & CCGA   & $-12,166.21$ & $178.23$ & $-12,663.12$ & $-11,850.92$ & \multirow{2}{*}{$0.0067$}\\ \cline{2-6}
$(-11,569.02)$ & K2     & $-12,226.24$ & $161.33$ & $-12,597.60$ & $-11,827.39$ & \\\hline\hline
Alarm 3000  & CCGA      & $-35,020.31$ & $396.06$ & $-36,275.13$ & $-34,093.27$ & \multirow{2}{*}{$0.0125$}\\ \cline{2-6}
$(-33,759.28)$ & K2     & $-35,138.41$ & $341.31$ & $-36,291.70$ & $-34,491.62$ & \\  \hline\hline
Alarm 5000  & CCGA      & $-57,282.66$ & $548.55$ & $-58,636.40$ & $-56,185.14$ & \multirow{2}{*}{$<0.0001$}\\ \cline{2-6}
$(-55,575.11)$ & K2     & $-57,574.12$ & $492.94$ & $-58,884.39$ & $-56,111.89$ & \\\hline 
\end{tabular}
\end{table*}

\begin{table*}
\centering
\caption{Results with the Insurance network.}
\begin{tabular}{|c|c|c|c|c|c|c|} \hline
\textbf{Data Set} & \textbf{Algorithm} & \textbf{Average} & \textbf{Standard Deviation} & \textbf{Minimum} & \textbf{Maximum} & \textbf{$p$-value}\\ \hline
Insurance 1000   & CCGA      & $-15,787.28$ & $279.19$ & $-16,581.80$ & $-15,180.42$ &  \multirow{2}{*}{$<0.0001$}\\ \cline{2-6}
$(-15,397.00)$ & K2      & $-16,107.61$ & $293.06$ & $-16,928.16$ & $-15,458.19$ & \\\hline\hline
Insurance 3000     & CCGA   & $-45,142.37$ & $722.67$ & $-47,530.71$ & $-43,789.77$ &  \multirow{2}{*}{$<0.0001$}\\ \cline{2-6}
$(-43,508.63)$ & K2     & $-45,778.39$ & $721.08$ & $-47,753.82$ & $-44,465.06$ & \\ \hline\hline
Insurance 5000  & CCGA      & $-74,624.88$ & $995.01$  & $-77,083.61$ & $-72,526.31$ &  \multirow{2}{*}{$<0.0001$}\\ \cline{2-6}
$(-72,183.51)$  & K2    & $-75,837.82$ & $1,249.83$& $-79,275.45$ & $-73,011.77$ & \\\hline 
\end{tabular}
\end{table*}

\begin{figure*}
\centering
\includegraphics[scale=0.53]{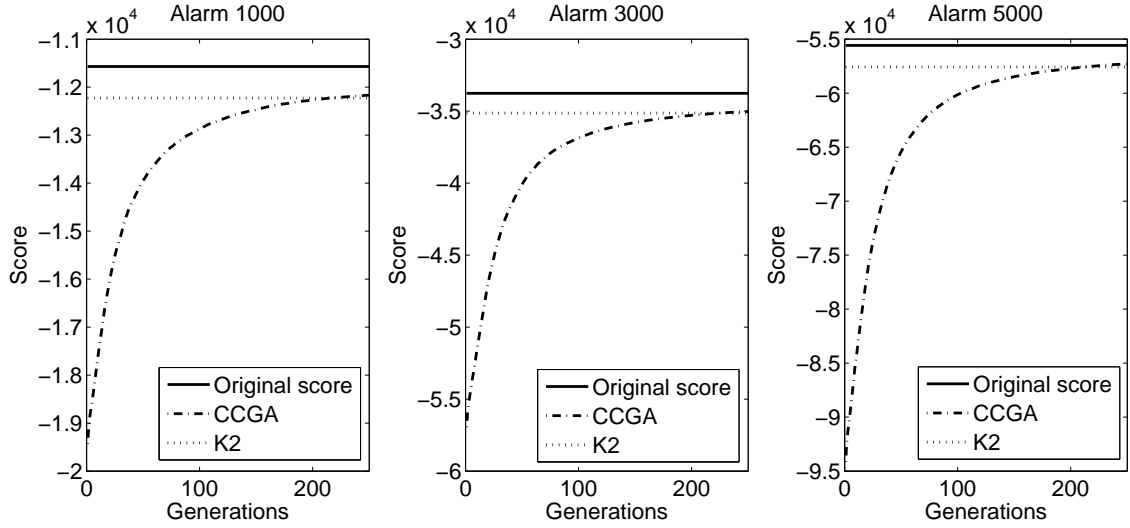}
\caption{Convergence of CCGA, averaged over 100 runs, on data sets from the Alarm network.}
\vskip -6pt
\end{figure*}

\begin{figure*}
\centering
\includegraphics[scale=0.53]{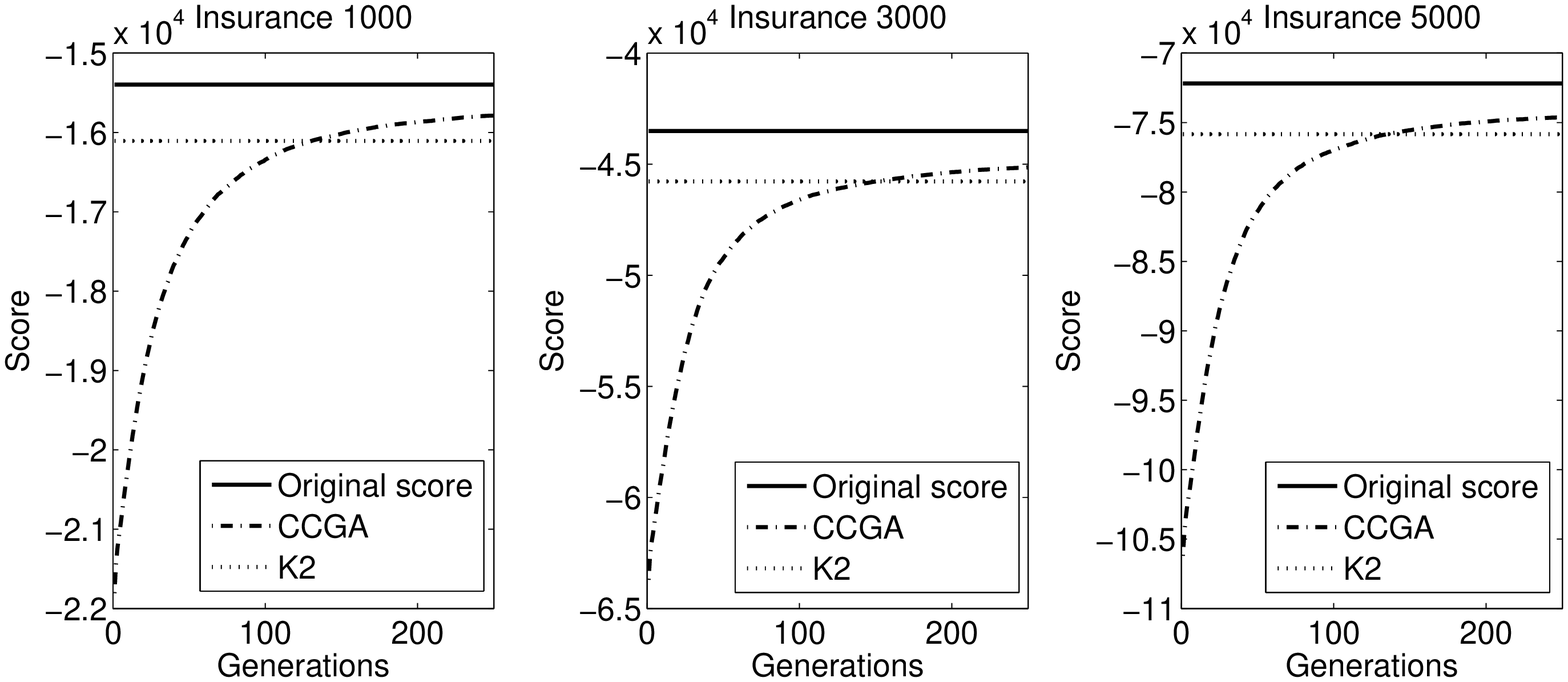}
\caption{Convergence of CCGA, averaged over 100 runs, on data sets from the Insurance network.}
\vskip -6pt
\end{figure*}

As can be seen from Table 3, CCGA slightly outperforms the K2 algorithm in all data sets from the Alarm network. Although some maximum and/or minimum values resulting from the K2 algorithm are higher, the average scores resulting from CCGA are greater than the average scores from K2. The $p$-values show that the differences between the average scores are statistically significant. From Table 4, we can see that the performance of CCGA on data sets from the Insurance network is even better. Both the average and the extreme scores from CCGA are greater than those resulting from  K2. Again, the $p$-values confirm that the differences between the average scores are statistically significant.

About the convergence of CCGA, Figure 5 shows that for data sets from the Alarm network, the average score resulting from CCGA becomes greater than the average score resulting from the K2 algorithm after, approximately, 230 generations. This value is around 150 for data sets from the Insurance network (Figure 6). We can see the result of our elitism-preserving approach in the fact that the average scores resulting from CCGA are monotonically increasing.

About the computational time spent by the algorithms, K2 was usually two orders of magnitude faster than CCGA (measured in seconds). However, we note that there is a lot of room for optimization in our CCGA implementation. First, we do not exploit parallelism. CCGA is highly suitable for parallel computation since the subpopulations evolve independently of each other in each generation (lines 8 to 13, Algorithm 1). Furthermore, differently from the K2 algorithm, we do not exploit the score decomposability property of the BDe score. Since the best individual of each subpopulation is used to create several complete solutions, we expect a lot of them to be very similar to each other. Thus, we can improve the performance of CCGA by using local scores together with standard bookkeeping techniques. We leave these code optimizations as future work.

\section{Related Work}

Genetic algorithms have become a popular method for learning Bayesian network structures. One of the pioneering work was done by Larrañaga \textit{et al.}~\cite{Larranaga94structurelearning}. They propose a genetic algorithm where a DAG is represented by a connectivity matrix, which is stored as the concatenation of its columns. Since this approach violates the acyclicity constraint, the authors use a repair operator that randomly eliminates edges that produce cycles. The authors also use the proposed algorithm assuming a predefined ordering between the nodes, thus removing the necessity of the repair operator. Few years later, Larrañaga \textit{et al.}~\cite{K2_ga} propose a hybrid genetic algorithm that searches for an optimal ordering of the nodes that is passed on to the K2 algorithm. The authors study the behaviour of the proposed algorithm with respect to different combinations of crossover and mutation operators. However, it is not clear whether their approach outperforms the K2 algorithm with random orderings.


Cooperative coevolution has also been used for learning Bayesian network structures. Wong \textit{et al.}~\cite{Wong} propose a hybrid method that combines characteristics of constrained-based and score-based approaches. The proposed algorithm has two main phases. First, it performs conditional independence tests to reduce the size of the search space. Thereafter, it uses a cooperative coevolutionary genetic algorithm to find a near-optimal network structure in the reduced search space. This genetic algorithm uses a different representation than the one proposed in this paper. In detail, it divides the network learning problem of $n$ variables into $n$ sub-problems, the goal of which is to find the ``optimal" parent set for the underlying node. To avoid cycles in complete solutions, the authors propose a feedback mechanism that uses the node ordering implied by a complete solution $x$ to produce constraints for each subpopulation such that new complete solutions will conform with that ordering. $x$ is later updated with results from newer complete solutions. A major drawback with this approach is that, for any update, $x$ will conform to the same ordering. This is equivalent to use a fixed, predefined ordering. The authors propose a solution to this issue, which is to associate every directed edge with a degree of belief. When the current degree of belief is less than a fixed threshold, the \emph{belief factor}, this suggests that the ordering imposed by the underlying edge may be wrong. Hence, a new ordering different than the original one can be used. We note that this solution brings back the original problem, namely the existence of cycles.

We can see our algorithm as an evolution of the previous genetic algorithms for learning Bayesian network structures from fully observable data sets. First, it does not require a predefined ordering of nodes because it coevolves multiple orderings. Furthermore, connectivity matrices are represented by strictly upper triangular matrices, thus ensuring that complete solutions do not violate the acyclicity constraint. Consequently, repair operators are not necessary.

\section{Conclusion}

We proposed a cooperative coevolutionary genetic algorithm for learning Bayesian network structures from fully observable data sets. Our proposed representation exploits the fact that this learning problem can be decomposed into two dependent subtasks, that is to find an optimal ordering of the nodes and an optimal connectivity matrix. We compared the performance of our solution with the deterministic algorithm K2 using six data sets generated from two traditional Bayesian networks, namely the Alarm network and the Insurance network. The results showed that our solution obtained better average scores for all data sets.

There are several exciting directions for future research work. First, we note that our algorithm does not restrict the number of parents that a node may have. However, the number of entries in a conditional probability table grows exponentially with the number of parents of the underlying node. Thus, the statistical cost of adding a parent to a node can be very large. An interesting extension of our algorithm is to explicitly penalize complete solutions that have nodes with an excessive number parents. This would seriously improve the computational efficiency of CCGA.

Furthermore, we intend to extend CCGA to deal with partially observable data sets. The problem of learning Bayesian network structures from incomplete data is more difficult than learning them from fully observable data sets. The major issue is that the BDe score no longer exists in closed form since it involves sufficient statistics that are not known when the data are incomplete. An interesting approach to circumvent this problem is to use CCGA with a third subpopulation that evolves the missing values. This would allow us to use the original BDe score together with the power of CCGA to find good network structures.

Finally, our novel representation can be of value to other graph-related problems, \textit{e.g.}, the TSP. Thus, we intend to investigate extensions of our algorithm to these problems.

\bibliographystyle{abbrv}

\end{document}